\documentclass{article}

\usepackage[nonatbib, final]{nips_2017}

\pdfoutput=1

\usepackage[utf8]{inputenc} 
\usepackage[T1]{fontenc}    
\usepackage{hyperref}       
\usepackage{url}            
\usepackage{booktabs}       
\usepackage{amsfonts}       
\usepackage{nicefrac}       
\usepackage{microtype}      
\usepackage{graphicx}
\usepackage{color,soul}
\usepackage{amsmath}
\usepackage{subcaption}
\DeclareMathOperator*{\argmax}{argmax}

\title{A unifying Bayesian approach for preterm `brain-age' prediction that models EEG sleep transitions over age}

\author{
  Kirubin M.~Pillay
  \\
  Institute of Biomedical Engineering\\ 
  Department of Engineering  Science\\
  University of Oxford\\
  Oxford, United Kingdom \\
  \texttt{kirubin.pillay@eng.ox.ac.uk}\\
 \And Maarten De Vos
 \\
  Institute of Biomedical Engineering\\
  Department of Engineering Science\\
  University of Oxford\\
  Oxford, United Kingdom \\
  \texttt{maarten.devos@eng.ox.ac.uk}
}

\begin{document}

\maketitle

\begin{abstract}
Preterm newborns undergo various stresses that may materialize as learning problems at school-age. Sleep staging of the Electroencephalogram (EEG), followed by prediction of their `brain-age' from these sleep states can quantify deviations from normal brain development early (when compared to the known age). Current automation of this approach relies on explicit sleep state classification, optimizing algorithms using clinicians' visually labelled sleep stages, which remains a subjective gold-standard. Such models fail to perform consistently over a wide age range and impacts the subsequent brain-age estimates that could prevent identification of subtler developmental deviations. We introduce a Bayesian Network utilizing multiple Gaussian Mixture Models, as a novel, unified approach for directly estimating brain-age, simultaneously modelling for both age and sleep dependencies on the EEG, to improve the accuracy of prediction over a wider age range. \end{abstract}

\section{Introduction}
\vspace{-0.1cm}
A preterm baby, born before the expected term age of 37-40 weeks Postmenstrual Age (PMA, the age since the last menstrual cycle of the mother) is admitted to the Neonatal Intensive Care Unit (NICU) to be cared for until reaching term age. During this period, the brain continues to mature, reflected by the change in their sleep state properties, as seen by Electroencephalogram (EEG) recordings \cite{Graven2008, Graven2006}. Deviations in the rate of sleep development may be indicative of cognitive problems by school-age \cite{Aarnoudse-Moens2009,Back2014}. Visual labelling of EEG sleep stages by trained clinicians and comparison with the given PMA can identify maturational delays, but this remains tedious and infeasible in the NICU setting \cite{Grigg-Damberger2016}. 

Consequently, methods have aimed to automate this `clinically driven' approach. More recently, Dereymaeker, Pillay et al. \cite{Dereymaeker2017} developed an algorithm to classify the sleep stages of Quiet Sleep (QS) and non-Quiet Sleep (non-QS), based on K-means clustering. From these sleep stages, features are derived from the EEG that shows significant correlations with PMA, and combined using multivariate regression to estimate `brain-age'. Discrepencies between brain-age and PMA are identified as an early-warning indication of abnormality \cite{DeWel2017,OToole2016}. This strategy, however, relies on the visual labelling of sleep stages (for optimizing the sleep-staging algorithms) which remains subjective \cite{Grigg-Damberger2016}, and the entire procedure requires different algorithms to fulfil each step and obtain a final brain-age estimate. No current sleep-detection algorithm works optimally across the full preterm age range, with particular difficulty seen at ages <30 weeks \cite{Dereymaeker2017}. This compromises the subsequent regression estimates and limits the age range within which we are confident of detecting key deviations.  

We introduce a Bayesian Network (BN) model as a novel, unified approach for estimating brain-age from an EEG recording, with validation on a real preterm dataset. This method models the dependencies of the EEG on both sleep state and age and is the first approach to be directly optimized on age, in an attempt to accurately estimate brain-age over a wider range, without requiring a consistently optimal sleep-state classification performance through an intermediate algorithm. 
 
\section{An Age-dependent Bayesian Network}
\begin{figure}[h!] 
  \centering
  \includegraphics[width=\linewidth]{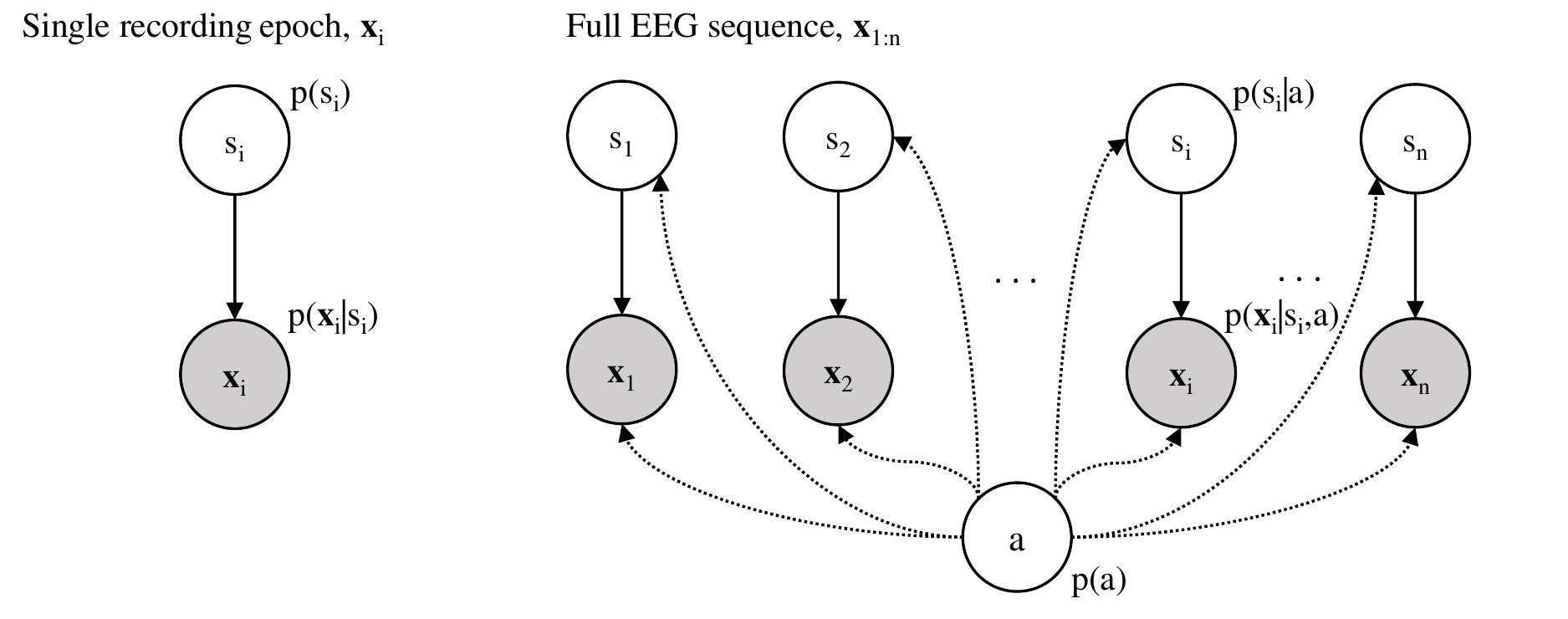}
  \caption{\textit{Left:} Graphical Bayesian Network (BN) representation to model the relationship between each $i^{th}$ EEG epoch, $\textbf{x}_i$,  and sleep state, $s_i$. \textit{Right:} BN extension to account for age dependency, $a$, across the entire recording, $\textbf{x}_{1:n}$.}
\label{bn}
\end{figure}

\vspace{-0.1cm}
Each EEG recording is divided into fixed-length epochs. Features are calculated from each $i^{th}$ epoch producing vectors denoted by the variable $\textbf{x}_i$ (of length $d$), with the full recording sequence given by $\textbf{x}_{1:n}$. Each epoch belongs to sleep states QS or non-QS, denoted by the variable $s_i$. Both states are consistently defined over age (although the underlying EEG properties continue to mature). We can model the dependency of $\textbf{x}_i$ on $s_i$ using a graphical BN representation, as shown in the left panel of Fig. \ref{bn}. With a BN, one can efficiently use the chain rule to represent the joint probability distribution across all variables from their conditional probabilities (taking into account conditional independencies) \cite{Russell1995}. The conditional probability for $\textbf{x}_i$ is $p(\textbf{x}_i|s_i)$, and $s_i$ is given by the prior $p(s_i)$. 

This BN can infer the sleep states from the recording at a single age. However, we are concerned with predicting the recording's age from an age-varying dataset. Denoting the age group by variable $a$, we expand the BN, as illustrated in the right panel of Fig. \ref{bn}. Sleep state probability is now conditioned on age, $p(s_i|a)$, as is each EEG epoch, $p(\textbf{x}_i|s_i,a)$. We choose to assume each epoch (and sleep state) as independent of one another, opting not to model dependencies between successive epochs. The extended BN can be described by the following hierarchical model:

\begin{align}
\textbf{x}_i|s_i,a &\sim f(a,s_i) \\
s_i|a &\sim \textrm{Bern}(q_a) \\
a &\sim \textrm{Cat}(\tfrac{1}{6}\textbf{1}_6) 
\end{align}

$a$ is a categorical distribution over 6 possible age groups, each of equal probability, as there is no physiological justification for one age being more likely than another. $s_i$ is  binary, sampled from a Bernoulli distribution with $q_a=p(s_i=QS|a)$. $f(a,s_i)$ is a set of Gaussian Mixture Models (GMMs), each dependent on both $a$ and $s_i$. For simplicity, the number of Gaussian components for each sleep state GMM, $k_{s_i}$, remains constant over age. In other words:
\begin{equation} 
f(a,s_i)=\sum_{j=1}^{k_{s_i}}c_{a,s_i,j}\mathcal{N}_{a,s_i,j}(\boldsymbol \mu_{a,s_i,j}, \boldsymbol \Sigma_{a,s_i,j})
\end{equation} 

$c_{a,s_i,j}$, $\boldsymbol \mu_{a,s_i,j}$ and $\boldsymbol \Sigma_{a,s_i,j}$ denote the Gaussian component weightings, means, and covariances, respectively. 

\section{Inference}
The aim is to predict the age given the recording, $p(a|\textbf{x}_{1:n})$. We begin by describing the posterior, $p(a|\textbf{x}_{1:n})$, using Bayes Rule. As we are concerned simply with the maximum posterior to classify age, we can drop the normalization term and approximate $p(a|\textbf{x}_{1:n})$ by:
\begin{equation} \label{Bayes}
p(a|\textbf{x}_{1:n}) = \frac{p(a)p(\textbf{x}_{1:n}|a)}{\sum_{a}p(a)p(\textbf{x}_{1:n}|a)} \propto p(a)p(\textbf{x}_{1},\textbf{x}_2,...,\textbf{x}_n|a)
\end{equation} 
  
From Fig. \ref{bn}, we note each epoch $\textbf{x}_i$ is conditionally independent of one another, given $a$. We can apply this assumption to Eqn. (\ref{Bayes}) treating the overall likelihood as a product of the individual epoch likelihoods, converting the result to a joint probability:
   
\begin{equation} 
p(a|\textbf{x}_{1:n}) \propto p(a)\prod_{i=1}^{n}p(\textbf{x}_i|a) = p(a)\prod_{i=1}^{n}\left[\frac{p(\textbf{x}_i,a)}{p(a)}\right]
\end{equation} 

We express $p(\textbf{x}_i,a)$ in terms of $s_i$ by marginalizing over $s_i$ and applying the chain rule to represent this in terms of the conditional probabilities that underpin the BN, before simplifying these terms:

\begin{equation} \label{post}
p(a|\textbf{x}_{1:n}) \propto p(a)\prod_{i=1}^{n}\left[ \frac{\sum_{s_i}p(\textbf{x}_i,s_i,a)}{p(a)}\right] = p(a)\prod_{i=1}^{n}\left[ \sum_{s_i}p(\textbf{x}_i|s_i,a)p(s_i|a)\right] 
\end{equation} 



To prevent underflow, we convert Eqn. (\ref{post}) to a log posterior, and obtain an estimate of the recording's age, $\hat{a}$, from the \textit{maximum a posteriori}:

\vspace{-0.2cm}
\begin{equation}\label{log}
\ln\left[ p(a|\textbf{x}_{1:n})\right] \propto \ln\left[p(a)\right] +  \sum_{i=1}^{n}\ln\left[\sum_{s_i}p(\textbf{x}_i|s_i,a)p(s_i|a)\right] \quad\mathrm{and}\quad  \hat{a}=\argmax_a \left\lbrace \ln\left[ p(a|\textbf{x}_{1:n})\right]\right\rbrace 
\end{equation} 
 

Using $\hat{a}$, we can derive a further inference equation from the model to additionally estimate the sleep state, $\hat{s_i}$, from $\textbf{x}_i$, if required: 

\begin{equation}
p(s_i|\textbf{x}_i,\hat{a})\propto p(\textbf{x}_i|s_i,\hat{a})p(s_i|\hat{a}) \quad\mathrm{and}\quad
\hat{s_i}=\argmax_{s_i} \left\lbrace p(s_i|\textbf{x}_i,\hat{a})\right\rbrace 
\end{equation}

\section{Experimental Methods}\label{exp}
\vspace{-0.1cm}
\subsection{Dataset and Features}
102 EEG recordings from 40 preterm patients were used in this study to validate the model, with the recordings aged between 27-42 weeks PMA and duration 1h40m-17h50m. All patients were clinically assessed as normal at 24 months of age, such that the brain age is expected to match the PMA. Recordings were organized into 6 age groups: <30 weeks (8 recordings), 30-31 weeks (17), 32-33 weeks (21), 34-35 weeks (14), 36-37 weeks (19), and $\geq$38 weeks (23). Each EEG recording was also visually labelled by clinicians for QS and non-QS, in 30s epochs. From each epoch, amplitude, spectral, and synchrony-based features were extracted. We identified the best features separating for both age and sleep state using Minimum Redundancy Maximum Relevance (mRMR) \cite{Vergara2014,Peng2005} by converting the age and sleep labels into a single set of 12 augmented labels.

\subsection{Training and Testing} 
A 10-fold cross validation (CV) was applied, divided by patient (to prevent recordings from the same patient in both test and train sets). mRMR was performed each time only on the training data. Within each fold, a further 18-fold patient CV was also performed on the training set to tune the parameters using a grid search. Three parameters required tuning: the number of selected features ($d$), and the number of QS and non-QS Gaussian components ($k_{s_i}$). $c_{a,s_i,j}$, $\boldsymbol \mu_{a,s_i,j}$ and $\boldsymbol \Sigma_{a,s_i,j}$, were determined by fitting GMMs to the training data for each age and sleep state group, using Expectation Maximisation \cite{McLachlinG;Peel2000}. $p(s_i|a)$ was also directly estimated from the training labels.  

We chose Krippendorff's Alpha for selecting optimal parameters and assessing final classification agreement between the model's brain-age estimates and known PMA \cite{Hayes2007}. Alpha accounts for chance (unlike accuracy) and can be calculated for both nominal ($\alpha_{nom}$) and ordinal ($\alpha_{ord}$) variables, of which $s_i$ is the former, and $a$ the latter (a misclassification near the true age group is considered better than one further away). Alpha values in the range 0.8-1 denotes an almost-perfect agreement \cite{Krippendorff2011}. 


\section{Results}
\begin{figure}[h!]
        \centering
        \begin{subfigure}[b]{\textwidth}
                \centering
                \includegraphics[width=0.5\textwidth]{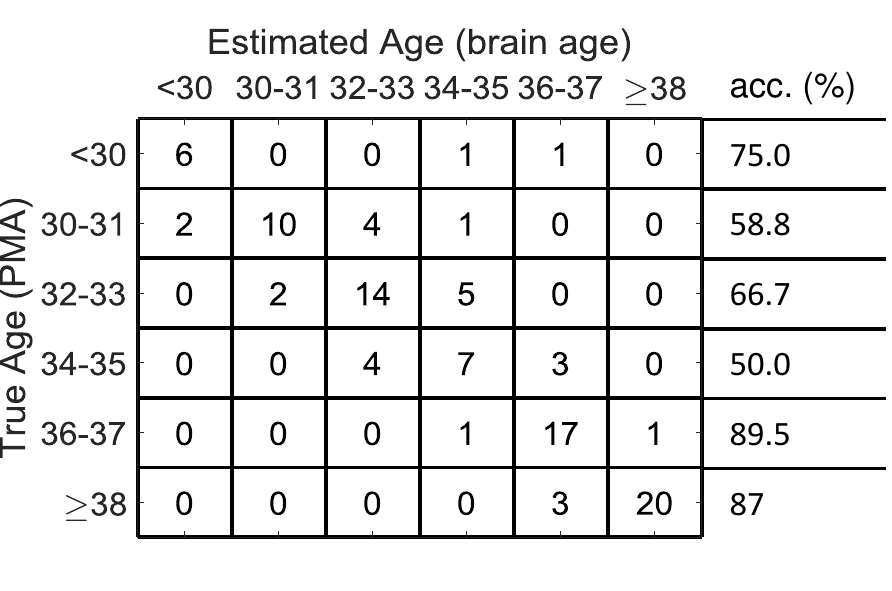}
                \vspace{-0.3cm}
                \caption{Confusion matrix and age-group classification accuracies (acc)}
                \label{conf_mat}
        \end{subfigure}%
        
        \begin{subfigure}[b]{\textwidth}
                \centering
                \includegraphics[width=\textwidth]{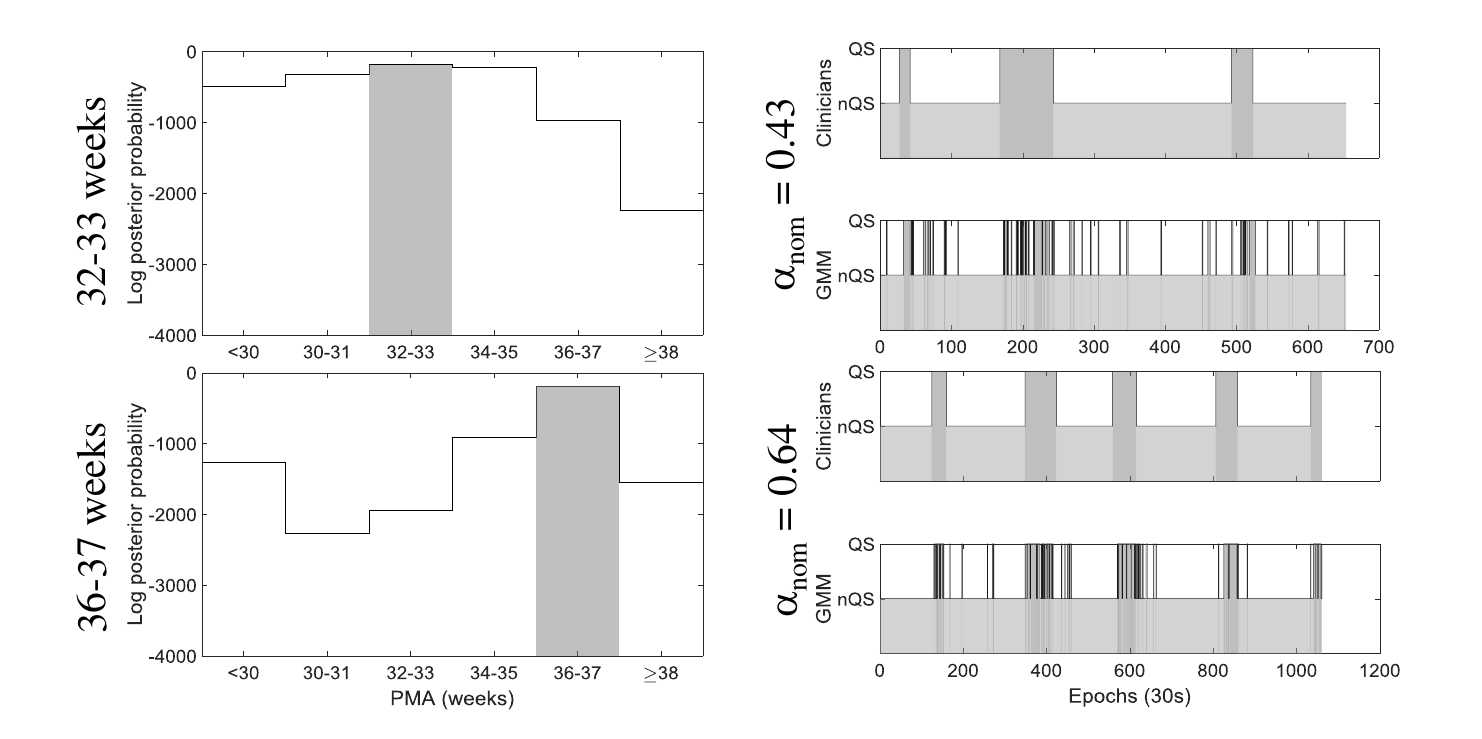}
                \caption{\textit{Left: }Posterior probabilities of two recordings at different ages (true age highlighted). \textit{Right:} Corresponding sleep state estimates by both clinicians and model, with alpha agreements ($\alpha_{norm}$).}
                \label{profiles}
        \end{subfigure}%
        \caption{Experimental results}\label{results}
\end{figure}

\vspace{-0.1cm}
Overall age classification accuracy was $72.5\%$ with corresponding $\alpha_{ord}$ of $0.91$. Fig. \ref{conf_mat} shows the confusion matrix, including classification accuracies (in \%) for each age group. The excellent value for $\alpha_{ord}$ is due to most misclassification occurring at adjacent age groups to the true value, due to some recordings existing near the defined group boundaries (e.g. a recording in group 34-35 weeks aged 35 weeks and 6 days is similar to one in group 36-37 weeks, aged 36 weeks). Only 3/102 recordings were misclassified beyond the adjacent ages. Fig. \ref{profiles} shows examples of posterior transitions over age for two recordings at 32-33 weeks and 36-37 weeks. Corresponding sleep state estimates by both the model and clinicians are also shown, with $\alpha_{nom}$ agreements.

\vspace{-0.1cm}

\vspace{-0.3cm}
\section{Discussion and Future Work}
\vspace{-0.1cm}
This is the first unified method for modelling sleep state transitions to directly and accurately predict brain-age, including in the difficult <30 week age group. Reduced data at certain ages remains a limitation, shown by the unusually high posterior estimate at <30 weeks for the 36-37 week recording (Fig. \ref{profiles}). We lay the groundwork for future uses of Bayesian graphical models in preterm outcome prediction, such as incorporating additional datasets (without visual labelling) using semi-supervised learning, expansion to a continuous age model, and further assessment on abnormal cohorts. 

\bibliographystyle{IEEEtran}
\bibliography{refs2}

\end{document}